\documentclass[conference]{IEEEtran}

\usepackage{cite}
\usepackage{amsmath,amssymb}
\usepackage{graphicx}
\usepackage{booktabs}
\usepackage{xcolor}
\usepackage{url}
\usepackage{tikz}
\usetikzlibrary{arrows.meta,positioning,fit,calc}

\newcommand{\qualref}{\IfFileExists{figures/qualitative.png}{ An example is shown in Fig.~\ref{fig:qual}.}{}}

\begin{document}
\bstctlcite{IEEEexample:BSTcontrol}

\title{Mirror Stress Test of TotalSegmentator and CADS: Position Dependence and Left--Right Labeling in CT Organ Segmentation}

\author{%
\IEEEauthorblockN{Ehsan Rassekh}
\IEEEauthorblockA{\textit{IEEE Graduate Student Member}\\
Email: rassekh.ehsan@gmail.com\textsuperscript{*}}
}

\maketitle
\begingroup\renewcommand{\thefootnote}{}\footnotetext{\textsuperscript{*}Academic email: rassekh.ehsan@spes.uniud.it}\endgroup

\begin{abstract}
Whole-body CT segmentation tools such as TotalSegmentator label more than one hundred structures, including pairs such as the left and the right kidney. Models of this kind are usually trained without left--right mirroring, because mirrored training images would make the left and right labels ambiguous. This work asks what such a model relies on: the appearance of an organ or its position in the image. A mirror stress test is proposed that needs no ground truth. Each CT scan is mirrored along its left--right axis, segmented, mirrored back and compared with the prediction on the original scan. TotalSegmentator is tested on 180 scans, and CADS, a second model trained by a different group on a much larger dataset, on 120 of the same scans, taken from three public Medical Segmentation Decathlon CT collections. For paired structures, the label follows the side of the image in both models: among the voxels found in both runs, the share carrying the swapped label is 1.000 for the kidneys and the lower lung lobes. For unpaired organs, the mean Dice between the two predictions is 0.00 for the spleen, the inferior vena cava and the gallbladder, and 0.33--0.40 for the liver, for both models and in all three collections. The two models fail on the same scans: the per-scan liver Dice of the two models has a correlation of 0.83. A full-resolution check and a confirmatory rule, fixed before the Pancreas run, lead to the same conclusion. Both models rely strongly on position, and their output should not be trusted on patients with mirrored anatomy without a check.
\end{abstract}


\section{Introduction}
Automatic segmentation of many anatomical structures in one pass has become a standard step in the analysis of CT images. Tools such as TotalSegmentator \cite{wasserthal2023totalsegmentator}, which is built on nnU-Net \cite{isensee2021nnunet}, label more than one hundred structures and are used for organ volumetry, opportunistic screening, radiotherapy planning and the preparation of research datasets. Many of these structures come in pairs: the left and the right kidney, adrenal gland, lung lobe, rib, hip bone and so on. A segmentation model must therefore decide not only \emph{which} structure is present but also \emph{on which side} it lies.

Mirroring is one of the most common data augmentations in segmentation, and nnU-Net uses it by default. For multi-organ CT it has a side effect: after a left--right mirror, a left kidney looks like a right kidney, so the label no longer matches the image unless the labels of all paired structures are swapped as well. Several well-known pipelines therefore remove it. The authors of TotalSegmentator report that mirroring was removed from the augmentation because otherwise the model could not separate left and right structures \cite{wasserthal2023totalsegmentator}; the CLIP-driven universal model also avoids mirroring to prevent confusion between left and right organs \cite{liu2023clip}; and a CBCT study reports that disabling left--right mirroring improved the results \cite{isensee2024cbct}. This choice is sensible, but it has a consequence that is rarely measured. A model that has never seen a mirrored patient can use the position of an organ in the image as a strong cue, and it may rely on this cue more than on the appearance of the organ.

This question has practical meaning. In situs inversus totalis, the thoracic and abdominal organs form a mirror image of the usual anatomy, with a frequency of about 1 in 10\,000 \cite{eitler2022situs}. Patients with this condition are also scanned, and their images may enter automatic pipelines. In such a case, a model that relies on position would look in the wrong place for the liver, the spleen and the stomach.

This work studies the question with a simple \emph{mirror stress test}. A CT scan is mirrored along its left--right axis, the model segments both the original and the mirrored scan, the second prediction is mirrored back, and the two predictions are compared. The test needs no ground-truth masks; it measures consistency, not accuracy. TotalSegmentator is used as the main model under test, CADS \cite{xu2026cads} is added as a second model, and public CT scans from the Medical Segmentation Decathlon \cite{antonelli2022msd} are used as input. The contributions of this work are:
\begin{itemize}
  \item A ground-truth-free mirror stress test with metrics that separate three behaviors for paired structures (label follows the anatomy, label follows the side of the image, or mixed) and that measure the loss of structures that are not paired.
  \item Evidence from 180 scans in three collections that the labels of paired structures follow the side of the image, and that the spleen, the inferior vena cava and the gallbladder are not found at all on mirrored scans, while the liver is found only in part.
  \item A second model, CADS, that shows the same behavior on 120 of the same scans, with failures on the same scans, and two differences: the heart chambers and the portal vein.
  \item A report of the decision rules used in this study, including a first rule that proved to be confounded and a confirmatory rule fixed before the last run.
\end{itemize}

\section{Related Work}
\subsection{Semantic Segmentation in General}
Semantic segmentation assigns a class to every pixel or voxel of an image. Fully convolutional networks \cite{long2015fcn} made dense prediction practical, and encoder--decoder designs such as U-Net \cite{ronneberger2015unet}, its volumetric version \cite{cicek20163dunet} and DeepLabv3+ \cite{chen2018deeplabv3plus} became the base of most later systems, in medical imaging as well as in road scenes. Earlier studies by the author compare such networks for scene understanding in autonomous vehicles \cite{rassekh2025semantic} and benchmark encoder backbones inside encoder--decoder models on driving scenes \cite{rassekh2026backbone}. A lesson that carries over to CT is that the result depends strongly on the design of the network and of the training procedure, and that a single score on held-out data does not show how a model behaves when the input changes. Transformer-based encoders, for example Swin UNETR with self-supervised pre-training \cite{tang2022swin}, and large benchmarks such as AMOS \cite{ji2022amos} have widened the range of tasks, but the usual evaluation is still the Dice score on scans with the usual anatomy.

\subsection{Multi-organ CT Segmentation}
nnU-Net \cite{isensee2021nnunet} showed that a carefully configured U-Net can reach very strong results on many segmentation tasks, and later work stressed the importance of rigorous validation and well-configured baselines \cite{isensee2024revisited}. TotalSegmentator \cite{wasserthal2023totalsegmentator} applies this recipe to whole-body CT with a large annotated dataset and has become a widely used tool. The CLIP-driven universal model \cite{liu2023clip} uses text embeddings of organ names to segment organs and tumors with one network. CADS \cite{xu2026cads} extends the idea to 167 structures from head to knee, with region-specific nnU-Net networks trained on 22,022 CT volumes that were collected from many sources and labeled with a semi-automatic pipeline. These models are mostly evaluated by the Dice score on held-out scans with the usual anatomy.

\subsection{Left--Right Handling in Segmentation Pipelines}
Left--right confusion is a known problem in multi-organ segmentation, but it is mostly handled as an engineering detail. A challenge submission based on nnU-Net reports that parts of the left kidney were sometimes labeled as the right kidney, and the same for the adrenal glands, and solves it with a connected-component post-processing step that assigns the side by position \cite{isensee2022extending}. A post-processing toolkit lists swapped left and right labels among the common errors of whole-body CT masks \cite{liu2025shapekit}. As mentioned above, mirroring is removed from training in several pipelines \cite{wasserthal2023totalsegmentator,liu2023clip,isensee2024cbct}. Together, these reports describe how the problem is avoided. To the best of the knowledge of this work, they do not measure what a model trained without mirroring does when it meets a mirrored scan.

\subsection{Evaluation Without Ground Truth}
Several studies assess segmentation models by comparing their own outputs under changes of the input. One study applies rotations and elastic deformations to CT scans, maps the predictions back and uses the agreement with the original prediction to estimate the quality of TotalSegmentator \cite{sayed2024evaluating}. Another compares several whole-body models by their concordance on the same scans \cite{giebeler2025concordance}. Test-time augmentation uses transformed inputs in a similar way, to estimate the uncertainty of a segmentation \cite{wang2019tta}. The mirror stress test follows the same logic, with a transformation that is deliberately anatomically unusual and that targets left--right behavior.

\section{Method}
\subsection{Mirror Stress Test}
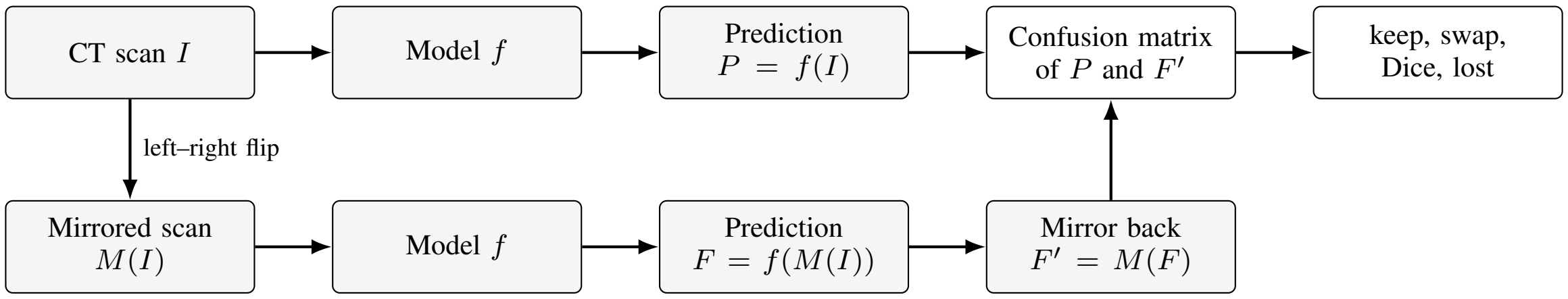
\begin{figure*}[t]
\centering
\begin{tikzpicture}[
  blk/.style={draw, rounded corners=2pt, align=center, minimum height=9mm, text width=2.2cm, font=\footnotesize, fill=black!4},
  arr/.style={-{Latex[length=2mm]}, thick}
]
\node[blk] (ct) at (0,0) {CT scan $I$};
\node[blk] (f1) at (3.2,0) {Model $f$};
\node[blk] (p) at (6.4,0) {Prediction\\$P=f(I)$};
\node[blk] (mi) at (0,-1.9) {Mirrored scan\\$M(I)$};
\node[blk] (f2) at (3.2,-1.9) {Model $f$};
\node[blk] (f) at (6.4,-1.9) {Prediction\\$F=f(M(I))$};
\node[blk] (mb) at (9.6,-1.9) {Mirror back\\$F'=M(F)$};
\node[blk, fill=white] (cmp) at (9.6,0) {Confusion matrix\\of $P$ and $F'$};
\node[blk, fill=white] (met) at (12.8,0) {keep, swap,\\Dice, lost};
\draw[arr] (ct) -- (f1);
\draw[arr] (f1) -- (p);
\draw[arr] (ct) -- node[right, font=\scriptsize]{left--right flip} (mi);
\draw[arr] (mi) -- (f2);
\draw[arr] (f2) -- (f);
\draw[arr] (f) -- (mb);
\draw[arr] (p) -- (cmp);
\draw[arr] (mb) -- (cmp);
\draw[arr] (cmp) -- (met);
\end{tikzpicture}
\caption{Mirror stress test. The model segments the original and the mirrored scan, the second prediction is mirrored back, and the two label maps are compared voxel by voxel. No ground truth is used.}
\label{fig:pipeline}
\end{figure*}

Fig.~\ref{fig:pipeline} shows the procedure. Let $I$ be a CT volume and $M$ the mirror operation along the left--right axis of $I$. The axis is found from the orientation codes of the image affine, and the array is flipped along it while the affine is kept unchanged, so that the result is the image of a mirrored patient. For a segmentation model $f$, the prediction on the original scan is $P=f(I)$, the prediction on the mirrored scan is $F=f(M(I))$, and the mirrored-back prediction is $F'=M(F)$. If the model was equivariant in an anatomical sense, $F'$ would show the same organs at the same places as $P$.

\subsection{Metrics}
Let $C_{p,q}$ be the number of voxels with label $p$ in $P$ and label $q$ in $F'$, counted over all voxels that are foreground in at least one of the two maps. For a class $c$, the Dice score between the two predictions without any label swap is
\begin{equation}
D_c=\frac{2\,C_{c,c}}{\sum_q C_{c,q}+\sum_p C_{p,c}}.
\end{equation}
For a pair of classes with left label $a$ and right label $b$, let $n_P=\sum_q (C_{a,q}+C_{b,q})$ be the size of the pair in $P$. The number of voxels that keep their label and the number of voxels that receive the partner label are
\begin{equation}
\mathrm{keep}=C_{a,a}+C_{b,b},\qquad \mathrm{swap}=C_{a,b}+C_{b,a}.
\end{equation}
Three quantities are derived from them. The share of side-swapped voxels among the voxels that are found in both runs is $\mathrm{swap}_{\mathrm{cond}}=\mathrm{swap}/(\mathrm{keep}+\mathrm{swap})$. The share of voxels of $P$ that are not found as the same pair in $F'$ is $\mathrm{lost}=1-(\mathrm{keep}+\mathrm{swap})/n_P$. The overlap of the pair without regard to the side is the Dice score of the two unions of left and right voxels. A pair is called \emph{anatomy-driven} in one case if $\mathrm{keep}/n_P\ge 0.9$, \emph{position-driven} if $\mathrm{swap}/n_P\ge 0.9$, and \emph{mixed} otherwise. A model that labels by the side of the image gives $\mathrm{swap}_{\mathrm{cond}}\approx 1$; a model that labels by the appearance of the organ gives $\mathrm{keep}/n_P\approx 1$. Structures from families such as the 12 ribs are pooled by summing the counts.

For unpaired structures there is no partner label, and $D_c$ is the only metric. A value near one means that the model finds the organ at the same place on the mirrored scan; a value near zero means that the organ is not found.

Two points should be noted. First, $P$ is not a ground truth, so a low $D_c$ can come from an error in $P$ as well as in $F'$; the test measures consistency, and the comparison across structures is therefore the informative part. Second, for symmetric paired structures such as the kidneys, the names ``left'' and ``right'' refer to the side of the body, so position-driven labeling is not an error in itself. It is informative because it shows what the model uses.

\subsection{Control and Decision Rules}
A control was run on the first two scans of every collection: the model was applied twice to the same original scan, and the two predictions agreed exactly (keep share 1.000), so the inference is deterministic in these cases.

The first decision rule, written before the first run, counted a structure family as problematic when its share of mixed cases was at least 0.20, and declared the audit positive when at least three eligible families (seen in at least 40 scans) were problematic. When it was applied to the first results, it turned out to be confounded: the category ``mixed'' contained cases where voxels were simply lost, such as for the small adrenal glands, and not cases where the sides were confused. The rule is reported as it was, together with the quantity $\mathrm{lost}$ that exposes the problem. A second analysis, the per-class Dice of unpaired organs, was added after seeing the first two collections and is therefore treated as exploratory. A confirmatory rule was then fixed in writing before the Pancreas run: the finding is confirmed if the median $D_c$ is below 0.1 for the spleen and below 0.5 for the liver on at least 40 scans of a collection not used before. The rule was not registered publicly.

\section{Experimental Setup}
\subsection{Models}
The main model under test is TotalSegmentator \cite{wasserthal2023totalsegmentator} in version~2.18.0, with the task \texttt{total} and its 117 classes. Pairs of left and right classes are found automatically from the class names, which gives 32 pairs. The main experiments use the fast variant, which works at 3\,mm resolution. A check on one collection uses the full-resolution variant.

The second model is CADS \cite{xu2026cads}, used with its released research-license weights (version~1.0.0) and code (version~1.0.3). It is a set of nnU-Net networks with the residual-encoder configuration, one for each anatomical group, and it works at the resolution of 1.5\,mm. Two of its groups are used: one with the abdominal organs, the lungs and the abdominal vessels (17 structures), and one with the esophagus, the heart chambers, the iliac vessels, the small bowel, the duodenum, the colon and the bladder (18 structures). Pairs of left and right structures are again found from the names. The identifier of the released weights shows that the networks were trained with a trainer without mirroring.

\subsection{Data}
Three CT collections of the Medical Segmentation Decathlon \cite{antonelli2022msd} are used: Hepatic Vessel, Colon and Pancreas. All three were collected at one institution and contain contrast-enhanced abdominal CT scans of patients with tumors. All data are public. Scans were drawn at random with a fixed seed: 100 scans each from Hepatic Vessel and Colon for a first run that gave the pair metrics, and 60 scans each from Hepatic Vessel, Colon and Pancreas for the main analysis that also stores the per-class Dice. The tables of this paper use the 60-scan runs; the 100-scan runs gave the same decisions for the first rule. A pair is analyzed only if both sides have at least 500 voxels in $P$, and a family is eligible if it appears in at least 40 scans. For the full-resolution check, 20 randomly drawn scans of the Colon collection were used. CADS was applied to 40 of the 60 scans of each collection (the first 40 in the order of the TotalSegmentator run, so that the scans are the same for both models), which gives 120 scans. In the Pancreas collection, two scans (case numbers 290 and 291) gave identical results in both models, which indicates a duplicated image in the source data. Both were kept; removing one of them changes no mean Dice value in this paper by more than 0.008.

\subsection{Statistics}
For each structure and collection, the mean of the per-scan Dice is reported with a 95\% percentile bootstrap interval (5000 resamples of the scans, fixed seed). The scans are the unit of resampling. No test of significance is used, because the main effects are large and the comparison is descriptive.

\subsection{Implementation}
The software reads each scan, mirrors it along the left--right axis found from the affine, runs the model on both scans, and stores the counts of the confusion matrix for every scan, so that a run can be resumed. The experiments ran on one NVIDIA L4 GPU with 24\,GB of memory. One scan took about 20--35\,s in the fast variant of TotalSegmentator and about 160\,s in its full-resolution variant, including both predictions. For CADS, one scan took about 90\,s for the two groups and both predictions, in batches of six scans.

\section{Results}
\subsection{Paired Structures Follow the Side of the Image}
Table~\ref{tab:pairs} gives the results for the main paired families. For the kidneys, the hip bones, the lower lobes of the lungs, the ribs, the iliopsoas and the autochthonous muscles, $\mathrm{swap}_{\mathrm{cond}}$ is 1.000 in all three collections, and the share of lost voxels is small (0.015--0.082). No family reached the anatomy-driven behavior: the keep share is at most 0.001 for all families in Table~\ref{tab:pairs}, except for the iliac arteries (0.03--0.10). The model finds the same structures at the mirrored place, and it gives them the label of the side where they now are. The behavior is the same for the lung lobes, where the left and right lung differ in shape: the lower lobes follow the side of the image.

\begin{table*}[t]
\caption{Paired structures on mirrored scans (60 random scans per collection). H: Hepatic Vessel, C: Colon, P: Pancreas. $n$ is the number of scans in which both sides are large enough. Swapped share is $\mathrm{swap}_{\mathrm{cond}}$; lost is the share of voxels of $P$ not found as the same pair in $F'$. In Hepatic Vessel, the hip family has fewer than 40 scans and the iliac artery family has 20.}
\label{tab:pairs}
\centering
\small
\begin{tabular}{@{}lccccccccc@{}}
\toprule
 & \multicolumn{3}{c}{$n$} & \multicolumn{3}{c}{Swapped share} & \multicolumn{3}{c}{Lost} \\
\cmidrule(lr){2-4}\cmidrule(lr){5-7}\cmidrule(lr){8-10}
Structure & H & C & P & H & C & P & H & C & P \\
\midrule
Kidney & 59 & 57 & 60 & 1.000 & 1.000 & 1.000 & 0.017 & 0.015 & 0.018 \\
Hip bone & 23 & 59 & 48 & 1.000 & 1.000 & 1.000 & 0.023 & 0.023 & 0.021 \\
Lung, lower lobe & 60 & 57 & 59 & 1.000 & 1.000 & 1.000 & 0.015 & 0.017 & 0.015 \\
Ribs & 60 & 57 & 60 & 1.000 & 1.000 & 1.000 & 0.069 & 0.040 & 0.047 \\
Iliopsoas & 60 & 60 & 60 & 1.000 & 1.000 & 1.000 & 0.082 & 0.081 & 0.081 \\
Autochthonous muscle & 60 & 60 & 60 & 1.000 & 1.000 & 1.000 & 0.057 & 0.060 & 0.058 \\
Adrenal gland & 57 & 52 & 58 & 0.999 & 0.998 & 1.000 & 0.548 & 0.584 & 0.536 \\
Iliac artery & 20 & 58 & 40 & 0.866 & 0.970 & 0.904 & 0.468 & 0.226 & 0.417 \\
\bottomrule
\end{tabular}
\end{table*}

Two groups differ from the rest. The adrenal glands have a swapped share of at least 0.998, but more than half of their voxels are lost (0.536--0.584), which is expected for small structures that are hard to segment even on the original scan. The iliac arteries and veins are lost in a large part (0.19--0.57) and show a lower swapped share; for the iliac veins it is 0.44 in Hepatic Vessel (12 scans) and 0.80 in Pancreas (27 scans), below the eligibility limit, which suggests that vessels may be handled partly by appearance. This was not studied further.

\IfFileExists{figures/qualitative.png}{%
\begin{figure*}[!t]
\centering
\includegraphics[width=\textwidth]{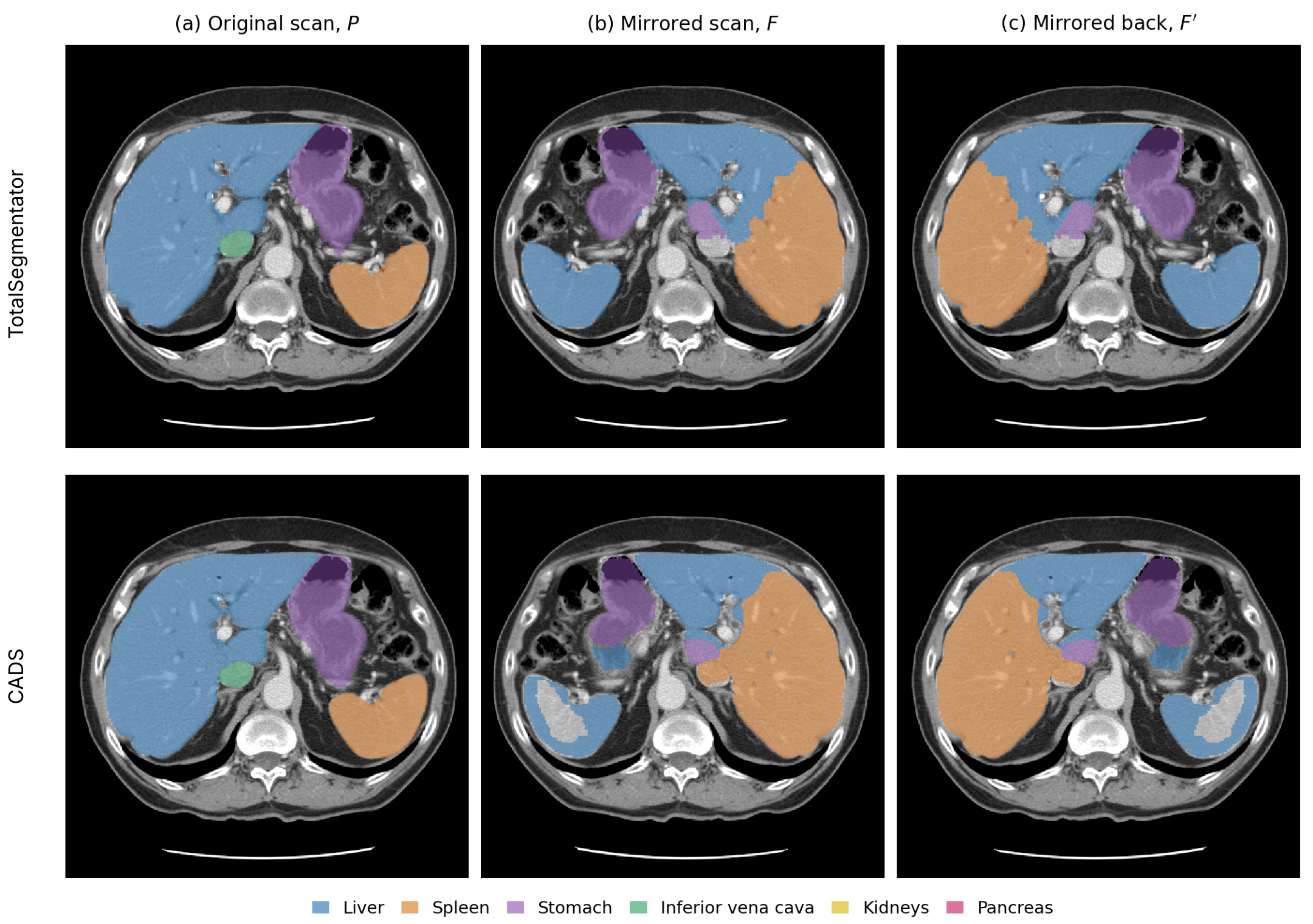}
\caption{Axial slice of one scan (pancreas\_001), segmented by TotalSegmentator (top row) and CADS (bottom row). (a) Prediction on the original scan. (b) Prediction on the mirrored scan. (c) The prediction of (b), mirrored back. In both models the spleen label lies over the liver and the liver label over the spleen in (c), and the inferior vena cava is not found.}
\label{fig:qual}
\end{figure*}}{}

\subsection{Unpaired Organs Are Lost on Mirrored Scans}
\begin{figure}[t]
\centering
\includegraphics[width=\columnwidth]{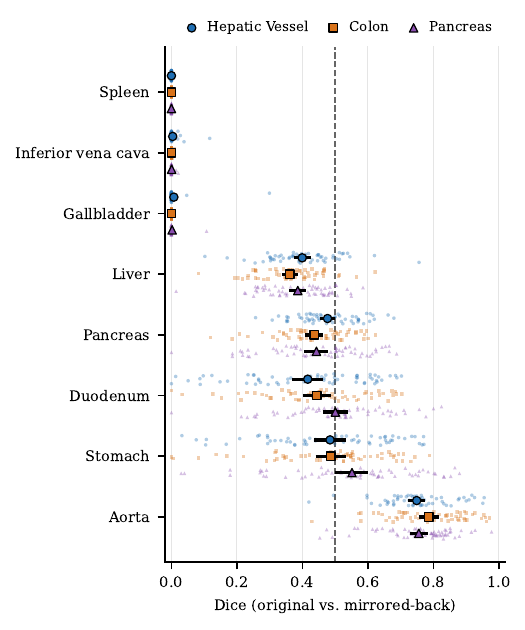}
\caption{Dice between the original and the mirrored-back prediction for each scan (small points) in the three collections. Large markers show the mean, and the black bars show the 95\% bootstrap interval of the mean. The dashed line marks a Dice of 0.5.}
\label{fig:cases}
\end{figure}

Table~\ref{tab:unpaired} shows the Dice score of unpaired structures between the original and the mirrored-back prediction, and Fig.~\ref{fig:cases} and Table~\ref{tab:ci} give the per-scan values and the intervals.\qualref{} The spleen, the inferior vena cava and the gallbladder are not found on mirrored scans: the mean Dice is 0.000--0.007 in all three collections, and in every scan it is below 0.5. The liver is found only in part (mean 0.361--0.399; Dice below 0.5 in 85\% to 93\% of the scans). The pancreas, the duodenum and the stomach are intermediate (mean 0.416--0.551), with 35\% to 72\% of the scans below 0.5. The aorta, the small bowel, the esophagus, the colon, the heart and the spinal cord are found in most scans, with mean values of 0.749--0.970. The values are very similar across the three collections. The intervals of the mean are narrow for the structures that are lost completely (the upper bound is 0.021 or less) and for the liver (width up to 0.053); they are wider for the duodenum and the stomach (up to 0.101). The per-scan points show why: the spleen, the inferior vena cava and the gallbladder form a single group at zero, while the liver, pancreas, duodenum and stomach are spread over most of the range between 0.2 and 0.8.

\begin{table*}[t]
\caption{Unpaired structures on mirrored scans: mean Dice between the original and the mirrored-back prediction, and the share of scans (\%) with a Dice below 0.5. H: Hepatic Vessel, C: Colon, P: Pancreas (60 random scans each; per structure $n$ = 47--60 in H, 53--60 in C and 49--60 in P).}
\label{tab:unpaired}
\centering
\small
\begin{tabular}{@{}lccccccc@{}}
\toprule
 & \multicolumn{3}{c}{Mean Dice} & \multicolumn{3}{c}{Scans with Dice $<0.5$ (\%)} \\
\cmidrule(lr){2-4}\cmidrule(lr){5-7}
Structure & H & C & P & H & C & P \\
\midrule
Spleen & 0.000 & 0.000 & 0.000 & 100 & 100 & 100 \\
Inferior vena cava & 0.004 & 0.001 & 0.001 & 100 & 100 & 100 \\
Gallbladder & 0.007 & 0.000 & 0.002 & 100 & 100 & 100 \\
Liver & 0.399 & 0.361 & 0.386 & 85.0 & 93.0 & 88.3 \\
Pancreas & 0.477 & 0.436 & 0.443 & 65.0 & 71.9 & 55.0 \\
Duodenum & 0.416 & 0.444 & 0.501 & 63.3 & 63.2 & 45.0 \\
Stomach & 0.484 & 0.487 & 0.551 & 46.7 & 40.4 & 35.0 \\
Portal and splenic vein & 0.628 & 0.545 & 0.653 & 11.7 & 31.6 & 11.7 \\
Aorta & 0.749 & 0.786 & 0.755 & 3.3 & 1.8 & 5.0 \\
Small bowel & 0.794 & 0.861 & 0.832 & 5.0 & 0.0 & 1.7 \\
Esophagus & 0.837 & 0.792 & 0.821 & 1.7 & 5.3 & 1.7 \\
Colon & 0.863 & 0.840 & 0.880 & 1.7 & 0.0 & 0.0 \\
Heart & 0.903 & 0.907 & 0.897 & 1.8 & 0.0 & 1.7 \\
Spinal cord & 0.970 & 0.967 & 0.970 & 0.0 & 0.0 & 0.0 \\
\bottomrule
\end{tabular}
\end{table*}

\begin{table*}[!t]
\caption{Mean Dice and 95\% bootstrap interval for unpaired structures on mirrored scans. H: Hepatic Vessel, C: Colon, P: Pancreas.}
\label{tab:ci}
\centering
\footnotesize
\setlength{\tabcolsep}{9pt}
\begin{tabular}{lccc}
\hline
Structure & H & C & P\\
\hline
Spleen & 0.000 [0.000, 0.000] & 0.000 [0.000, 0.000] & 0.000 [0.000, 0.000]\\
Inferior vena cava & 0.004 [0.001, 0.009] & 0.001 [0.000, 0.001] & 0.001 [0.000, 0.002]\\
Gallbladder & 0.007 [0.000, 0.021] & 0.000 [0.000, 0.000] & 0.002 [0.000, 0.007]\\
Liver & 0.399 [0.373, 0.426] & 0.361 [0.338, 0.386] & 0.386 [0.359, 0.412]\\
Pancreas & 0.477 [0.455, 0.499] & 0.436 [0.409, 0.464] & 0.443 [0.405, 0.479]\\
Duodenum & 0.416 [0.367, 0.464] & 0.444 [0.400, 0.486] & 0.501 [0.462, 0.539]\\
Stomach & 0.484 [0.436, 0.533] & 0.487 [0.440, 0.534] & 0.551 [0.500, 0.601]\\
Aorta & 0.749 [0.724, 0.774] & 0.786 [0.755, 0.816] & 0.755 [0.727, 0.783]\\
\hline
\end{tabular}
\end{table*}

\subsection{Confirmatory Test}
The confirmatory rule was applied to the Pancreas collection, which was not used before. The median Dice is 0.000 for the spleen (limit 0.1) and 0.398 for the liver (limit 0.5), with 60 scans each. The rule is therefore satisfied. The first rule, applied literally, gave ``weak'' for Hepatic Vessel and Pancreas and ``positive'' for Colon, but its category ``mixed'' is dominated by lost voxels in the adrenal glands and the iliac vessels, as described above, and these verdicts are not used for the conclusion.

\subsection{Full-Resolution Check}
To test whether the result depends on the fast 3\,mm variant, the full-resolution variant was applied to 20 randomly drawn scans of the Colon collection (19 scans with a structure large enough; 18 for the gallbladder). Table~\ref{tab:full} gives the results. The pairs behave in the same way ($\mathrm{swap}_{\mathrm{cond}}$ is 1.000 for the kidneys, hips, ribs, lower lung lobes and iliopsoas). The spleen, the inferior vena cava and the gallbladder are still not found. The liver is found even less often (median 0.294), while the aorta, the pancreas and the heart improve. The two variants were run on different random subsets, so the numbers are not paired.

\begin{table}[t]
\caption{Fast (3\,mm, up to 57 scans) and full-resolution (up to 19 scans) variants on the Colon collection: Dice between the original and the mirrored-back prediction.}
\label{tab:full}
\centering
\small
\begin{tabular}{@{}lccc@{}}
\toprule
 & Fast & \multicolumn{2}{c}{Full resolution} \\
\cmidrule(lr){2-2}\cmidrule(lr){3-4}
Structure & Mean & Mean & Median \\
\midrule
Spleen & 0.000 & 0.000 & 0.000 \\
Inferior vena cava & 0.001 & 0.013 & 0.000 \\
Gallbladder & 0.000 & 0.002 & 0.000 \\
Liver & 0.361 & 0.267 & 0.294 \\
Duodenum & 0.444 & 0.449 & 0.480 \\
Stomach & 0.487 & 0.454 & 0.525 \\
Pancreas & 0.436 & 0.568 & 0.605 \\
Aorta & 0.786 & 0.891 & 0.932 \\
Heart & 0.907 & 0.942 & 0.946 \\
\bottomrule
\end{tabular}
\end{table}

\subsection{A Second Model: CADS}
CADS was tested on 120 scans, 40 per collection, with the same procedure. Table~\ref{tab:cadspairs} gives the paired structures. The result is the same as for TotalSegmentator: for the kidneys and the lower lobes of the lungs, $\mathrm{swap}_{\mathrm{cond}}$ is 1.000 in all three collections, with 1.2\% to 3.3\% of the voxels lost, and the adrenal glands reach 1.000 with about 64\% of the voxels lost. The iliac vessels give a swapped share of 0.97--0.99 in the Colon collection and lower values with more lost voxels in the other two. The upper lobes of the lungs are small on these abdominal scans and mostly lost (0.32--0.98). Again, no family with enough voxels follows the anatomy.

The heart chambers, which are not part of the \texttt{total} task of TotalSegmentator, show a different pattern. The swapped share is only 0.02--0.08, so the voxels that are found keep the label of the chamber and not the label of the side, but 62\% to 77\% of the voxels are lost, and the myocardium is not found (mean Dice 0.005--0.010, below 0.5 in all scans). For the heart, the model therefore uses the anatomy for the voxels that it finds, but it does not find most of them.

\begin{table*}[t]
\caption{CADS: paired structures on mirrored scans (40 scans per collection). H: Hepatic Vessel, C: Colon, P: Pancreas. $n$ is the number of scans in which both sides have at least 500 voxels; the counts are summed over the scans.}
\label{tab:cadspairs}
\centering
\small
\begin{tabular}{@{}lccccccccc@{}}
\toprule
 & \multicolumn{3}{c}{$n$} & \multicolumn{3}{c}{Swapped share} & \multicolumn{3}{c}{Lost} \\
\cmidrule(lr){2-4}\cmidrule(lr){5-7}\cmidrule(lr){8-10}
Structure & H & C & P & H & C & P & H & C & P \\
\midrule
Kidney & 40 & 38 & 40 & 1.000 & 1.000 & 1.000 & 0.021 & 0.018 & 0.033 \\
Lung, lower lobe & 40 & 38 & 39 & 1.000 & 1.000 & 1.000 & 0.016 & 0.012 & 0.013 \\
Lung, upper lobe & 39 & 38 & 38 & 1.000 & 1.000 & 0.999 & 0.983 & 0.320 & 0.908 \\
Adrenal gland & 40 & 38 & 40 & 1.000 & 1.000 & 1.000 & 0.636 & 0.636 & 0.639 \\
Iliac artery & 19 & 40 & 35 & 0.868 & 0.992 & 0.940 & 0.510 & 0.211 & 0.423 \\
Iliac vein & 9 & 39 & 19 & 0.492 & 0.966 & 0.632 & 0.593 & 0.197 & 0.578 \\
Heart, atrium & 37 & 38 & 39 & 0.042 & 0.029 & 0.021 & 0.767 & 0.745 & 0.728 \\
Heart, ventricle & 39 & 38 & 39 & 0.078 & 0.077 & 0.058 & 0.616 & 0.652 & 0.651 \\
\bottomrule
\end{tabular}
\end{table*}

For the unpaired structures, Table~\ref{tab:cadsunpaired} compares the two models on the same scans. The values are close. The spleen, the inferior vena cava and the gallbladder are not found by either model (mean Dice at most 0.011), and the liver is found in part (CADS 0.327--0.394, TotalSegmentator 0.360--0.391; Dice below 0.5 in 92\% to 97\% of the scans for CADS, with 95\% bootstrap intervals of the mean that are at most 0.055 wide). The duodenum, the stomach, the aorta, the esophagus, the small bowel and the colon also have similar values. Two structures differ. The pancreas is more consistent in CADS (0.501--0.581 against 0.424--0.468), and the portal and splenic vein is much less consistent (0.212--0.274 against 0.528--0.656).

\begin{table*}[t]
\caption{Unpaired structures on mirrored scans for the two models on the same 40 scans of each collection: mean Dice between the original and the mirrored-back prediction. TS: TotalSegmentator, CADS: second model.}
\label{tab:cadsunpaired}
\centering
\small
\begin{tabular}{@{}lcccccc@{}}
\toprule
 & \multicolumn{2}{c}{Hepatic Vessel} & \multicolumn{2}{c}{Colon} & \multicolumn{2}{c}{Pancreas} \\
\cmidrule(lr){2-3}\cmidrule(lr){4-5}\cmidrule(lr){6-7}
Structure & TS & CADS & TS & CADS & TS & CADS \\
\midrule
Spleen & 0.000 & 0.000 & 0.000 & 0.000 & 0.000 & 0.000 \\
Inferior vena cava & 0.003 & 0.000 & 0.001 & 0.000 & 0.001 & 0.000 \\
Gallbladder & 0.011 & 0.003 & 0.000 & 0.000 & 0.003 & 0.000 \\
Liver & 0.391 & 0.394 & 0.360 & 0.327 & 0.389 & 0.375 \\
Pancreas & 0.468 & 0.581 & 0.437 & 0.501 & 0.424 & 0.580 \\
Duodenum & 0.408 & 0.484 & 0.431 & 0.463 & 0.483 & 0.528 \\
Stomach & 0.499 & 0.513 & 0.474 & 0.481 & 0.584 & 0.549 \\
Portal and splenic vein & 0.624 & 0.274 & 0.528 & 0.212 & 0.656 & 0.268 \\
Aorta & 0.767 & 0.741 & 0.786 & 0.755 & 0.776 & 0.744 \\
Esophagus & 0.843 & 0.858 & 0.804 & 0.812 & 0.822 & 0.810 \\
Small bowel & 0.802 & 0.822 & 0.862 & 0.873 & 0.823 & 0.816 \\
Colon & 0.862 & 0.901 & 0.847 & 0.876 & 0.882 & 0.906 \\
\bottomrule
\end{tabular}
\end{table*}

\begin{figure}[t]
\centering
\includegraphics[width=\columnwidth]{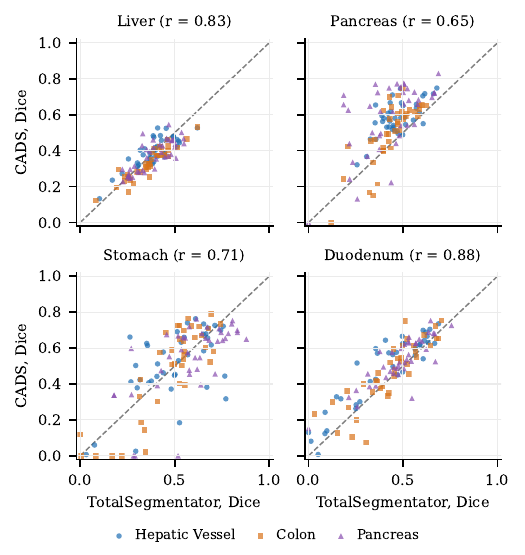}
\caption{Dice between the original and the mirrored-back prediction of the two models on the same scan, for four structures (118 scans with at least 500 voxels in $P$). The dashed line marks equal values, and $r$ is the Pearson correlation.}
\label{fig:agree}
\end{figure}

The two models also fail on the same scans. Fig.~\ref{fig:agree} shows the Dice of each model on the same scan. The correlation over the 118 scans in which the structure has at least 500 voxels in $P$ is 0.83 for the liver, 0.88 for the duodenum, 0.71 for the stomach and 0.65 for the pancreas. A scan in which the liver is found better by one model is found better by the other as well, which suggests that the difficulty depends on the scan and not on the model.

\section{Discussion}
\subsection{What the Model Uses}
Two observations point in the same direction. For paired structures, labels follow the side of the image: no family is anatomy-driven. For unpaired structures, the organs with a strong position prior, such as the spleen, the inferior vena cava and the gallbladder, are not found when they appear on the other side, and the liver is found only in part. The heart and the spinal cord, which are close to the midline, and large structures with a distinct shape such as the colon and the small bowel are found. A model trained on non-mirrored data can therefore use the position of an organ as a cue that is as strong as, or stronger than, its appearance. The second model supports this reading. CADS comes from a different group, was trained on a dataset that is about eighteen times larger than that of TotalSegmentator, with a different network configuration, but it gives the same pattern for paired structures and nearly the same values for the unpaired organs, and the two models fail on the same scans. The training data of the two models overlap, because CADS includes the TotalSegmentator scans with refined labels \cite{xu2026cads}. Studies of convolutional networks show that such networks can encode absolute position \cite{islam2020position,kayhan2020translation}, and the use of an easy cue instead of the intended one is known as shortcut learning \cite{geirhos2020shortcut}. One common factor that remains is the training without mirroring, so the result is consistent with the effect of this design choice, although a controlled training experiment would be needed to prove it. This agrees with the engineering reports that led to the removal of mirroring \cite{wasserthal2023totalsegmentator,liu2023clip,isensee2024cbct}, but to the knowledge of this work the size of the effect has not been measured before.

\subsection{Clinical Meaning}
In situs inversus totalis, the thoracic and abdominal organs are a mirror image of the usual anatomy \cite{eitler2022situs}. A mirrored scan is a simple model of this condition, and in this model the segmentation of the spleen and of the inferior vena cava fails completely and the segmentation of the liver fails in most scans. Volumes and masks that are derived from such predictions, for example for radiotherapy planning or volumetry, would be wrong or missing. Because the condition is rare, the practical risk is small, but the failure is silent: the model returns masks without any warning. A simple safeguard is to check the position of large organs, such as the liver and the spleen, before the masks are used.

\subsection{Limitations}
Several limits apply. First, the Dice score is computed against the prediction on the original scan, not against a manual annotation, so the test measures consistency. Second, a mirrored scan is an approximation of a patient with situs inversus: real patients may have additional anomalies, and the veins and arteries may follow different patterns. Third, two models were tested (TotalSegmentator, fast variant for most results, and CADS, on 40 scans per collection), and both were trained without mirroring and partly on the same scans; the behavior of models trained with mirroring is open. Fourth, the three collections come from one institution and include patients with tumors. Fifth, the adrenal glands and the iliac vessels are partly lost even on the usual orientation, so their results are hard to interpret. Sixth, the intervals in Table~\ref{tab:ci} are based on 57 to 60 scans per collection; they are wide for the structures with intermediate values, such as the stomach and the duodenum, and the scans of one collection are not independent of the patient population. Finally, part of the analysis was added after the first results, which is why the confirmatory rule was used.

\subsection{Possible Remedy}
The obvious remedy is label-aware mirroring: when a training image is mirrored, the labels of all left--right pairs are swapped. This was not tested here. Whether it removes the position dependence without hurting the usual cases is a question for training experiments, which are outside the scope of this work.

\section{Conclusion and Future Work}
This work proposed a mirror stress test for multi-organ CT segmentation models and applied it to TotalSegmentator on 180 public scans and to CADS on 120 of them. The labels of paired structures follow the side of the image, and several unpaired organs, in particular the spleen, the inferior vena cava and the gallbladder, are not found on mirrored scans, while the liver is found only in part. The result was confirmed on a collection that was not used before, it did not depend on the resolution of TotalSegmentator, and the second model gave the same pattern, with failures on the same scans. The models differ for the heart chambers, which keep the anatomical label when they are found, and for the portal vein, which is found less often by the second model.

Future work has three directions. First, other publicly available whole-body models, in particular models trained with mirroring, should be tested with the same procedure, to see how general the behavior is. Second, models should be trained with and without label-aware mirroring on the same data, to measure how much of the position dependence can be removed. Third, a small set of real CT scans of patients with situs inversus should be used to check how well the mirror stress test predicts the behavior on real mirrored anatomy.

\bibliographystyle{IEEEtran}
\bibliography{refs}

\begin{thebibliography}{10}
\providecommand{\url}[1]{#1}
\csname url@samestyle\endcsname
\providecommand{\newblock}{\relax}
\providecommand{\bibinfo}[2]{#2}
\providecommand{\BIBentrySTDinterwordspacing}{\spaceskip=0pt\relax}
\providecommand{\BIBentryALTinterwordstretchfactor}{4}
\providecommand{\BIBentryALTinterwordspacing}{\spaceskip=\fontdimen2\font plus
\BIBentryALTinterwordstretchfactor\fontdimen3\font minus
  \fontdimen4\font\relax}
\providecommand{\BIBforeignlanguage}[2]{{%
\expandafter\ifx\csname l@#1\endcsname\relax
\typeout{** WARNING: IEEEtran.bst: No hyphenation pattern has been}%
\typeout{** loaded for the language `#1'. Using the pattern for}%
\typeout{** the default language instead.}%
\else
\language=\csname l@#1\endcsname
\fi
#2}}
\providecommand{\BIBdecl}{\relax}
\BIBdecl

\bibitem{wasserthal2023totalsegmentator}
J.~Wasserthal, H.-C. Breit, M.~T. Meyer, M.~Pradella, D.~Hinck, A.~W. Sauter,
  T.~Heye, D.~T. Boll, J.~Cyriac, S.~Yang, M.~Bach, and M.~Segeroth,
  ``{TotalSegmentator}: Robust segmentation of 104 anatomic structures in {CT}
  images,'' \emph{Radiology: Artificial Intelligence}, vol.~5, no.~5, p.
  e230024, 2023.

\bibitem{isensee2021nnunet}
F.~Isensee, P.~F. Jaeger, S.~A.~A. Kohl, J.~Petersen, and K.~H. Maier-Hein,
  ``{nnU-Net}: a self-configuring method for deep learning-based biomedical
  image segmentation,'' \emph{Nature Methods}, vol.~18, no.~2, pp. 203--211,
  2021.

\bibitem{liu2023clip}
J.~Liu, Y.~Zhang, J.-N. Chen, J.~Xiao, Y.~Lu, B.~A. Landman, Y.~Yuan,
  A.~Yuille, Y.~Tang, and Z.~Zhou, ``{CLIP}-driven universal model for organ
  segmentation and tumor detection,'' in \emph{Proc. IEEE/CVF Int. Conf.
  Computer Vision (ICCV)}, 2023.

\bibitem{isensee2024cbct}
F.~Isensee, Y.~Kirchhoff, L.~Kraemer, M.~Rokuss, C.~Ulrich, and K.~H.
  Maier-Hein, ``Scaling {nnU-Net} for {CBCT} segmentation,'' arXiv:2411.17213,
  2024.

\bibitem{eitler2022situs}
K.~Eitler, A.~Bibok, and G.~Telkes, ``Situs inversus totalis: A clinical
  review,'' \emph{International Journal of General Medicine}, vol.~15, pp.
  2437--2449, 2022.

\bibitem{xu2026cads}
M.~Xu \emph{et~al.}, ``{CADS}: A comprehensive anatomical dataset and
  segmentation for whole-body anatomy in computed tomography,'' \emph{Nature
  Communications}, 2026.

\bibitem{antonelli2022msd}
M.~Antonelli, A.~Reinke, S.~Bakas \emph{et~al.}, ``The {Medical Segmentation
  Decathlon},'' \emph{Nature Communications}, vol.~13, p. 4128, 2022.

\bibitem{long2015fcn}
J.~Long, E.~Shelhamer, and T.~Darrell, ``Fully convolutional networks for
  semantic segmentation,'' in \emph{Proc. IEEE Conf. Computer Vision and
  Pattern Recognition (CVPR)}, 2015, pp. 3431--3440.

\bibitem{ronneberger2015unet}
O.~Ronneberger, P.~Fischer, and T.~Brox, ``{U-Net}: Convolutional networks for
  biomedical image segmentation,'' in \emph{Medical Image Computing and
  Computer-Assisted Intervention (MICCAI)}, ser. LNCS, vol. 9351, 2015, pp.
  234--241.

\bibitem{cicek20163dunet}
{\"O}.~{\c{C}}i{\c{c}}ek, A.~Abdulkadir, S.~S. Lienkamp, T.~Brox, and
  O.~Ronneberger, ``{3D U-Net}: Learning dense volumetric segmentation from
  sparse annotation,'' in \emph{Medical Image Computing and Computer-Assisted
  Intervention (MICCAI)}, ser. LNCS, vol. 9901, 2016, pp. 424--432.

\bibitem{chen2018deeplabv3plus}
L.-C. Chen, Y.~Zhu, G.~Papandreou, F.~Schroff, and H.~Adam, ``Encoder-decoder
  with atrous separable convolution for semantic image segmentation,'' in
  \emph{Proc. European Conf. Computer Vision (ECCV)}, 2018, pp. 801--818.

\bibitem{rassekh2025semantic}
E.~Rassekh, ``Semantic segmentation based scene understanding in autonomous
  vehicles,'' \emph{arXiv preprint arXiv:2507.14303}, 2025.

\bibitem{rassekh2026backbone}
E.~Rassekh, ``Backbone matters: Benchmarking encoder-decoder networks for
  semantic segmentation of driving scenes on {BDD100K},'' \emph{arXiv preprint
  arXiv:2611.26191}, 2026.

\bibitem{tang2022swin}
Y.~Tang, D.~Yang, W.~Li, H.~R. Roth, B.~Landman, D.~Xu, V.~Nath, and
  A.~Hatamizadeh, ``Self-supervised pre-training of {Swin} transformers for
  {3D} medical image analysis,'' in \emph{Proc. IEEE/CVF Conf. Computer Vision
  and Pattern Recognition (CVPR)}, 2022, pp. 20\,730--20\,740.

\bibitem{ji2022amos}
Y.~Ji \emph{et~al.}, ``{AMOS}: A large-scale abdominal multi-organ benchmark
  for versatile medical image segmentation,'' in \emph{Advances in Neural
  Information Processing Systems (NeurIPS), Datasets and Benchmarks Track},
  2022, arXiv:2206.08023.

\bibitem{isensee2024revisited}
F.~Isensee, T.~Wald, C.~Ulrich, M.~Baumgartner, S.~Roy, K.~Maier-Hein, and
  P.~J{\"a}ger, ``{nnU-Net} revisited: A call for rigorous validation in {3D}
  medical image segmentation,'' arXiv:2404.09556, 2024.

\bibitem{isensee2022extending}
F.~Isensee, C.~Ulrich, T.~Wald, and K.~Maier-Hein, ``Extending {nnU-Net} is all
  you need,'' arXiv:2208.10791, 2022.

\bibitem{liu2025shapekit}
J.~Liu, D.~He, W.~Li, N.~Wang, A.~L. Yuille, and Z.~Zhou, ``{ShapeKit},''
  arXiv:2506.24003, 2025.

\bibitem{sayed2024evaluating}
M.~Sayed, S.~Saba-Sadiya, B.~Wichtlhuber, J.~Dietz, M.~Neitzel, L.~Keller,
  G.~Roig, and A.~M. Bucher, ``Evaluating medical image segmentation models
  using augmentation,'' \emph{Tomography}, vol.~10, no.~12, pp. 2128--2143,
  2024.

\bibitem{giebeler2025concordance}
L.~Giebeler, D.~Krishnaswamy, D.~Clunie, J.~Wasserthal \emph{et~al.}, ``In
  search of truth: Evaluating concordance of {AI}-based anatomy segmentation
  models,'' arXiv:2512.15921, 2025.

\bibitem{wang2019tta}
G.~Wang, W.~Li, M.~Aertsen, J.~Deprest, S.~Ourselin, and T.~Vercauteren,
  ``Aleatoric uncertainty estimation with test-time augmentation for medical
  image segmentation with convolutional neural networks,''
  \emph{Neurocomputing}, vol. 338, pp. 34--45, 2019.

\bibitem{islam2020position}
M.~A. Islam, S.~Jia, and N.~D.~B. Bruce, ``How much position information do
  convolutional neural networks encode?'' in \emph{International Conference on
  Learning Representations (ICLR)}, 2020, arXiv:2001.08248.

\bibitem{kayhan2020translation}
O.~S. Kayhan and J.~C. van Gemert, ``On translation invariance in {CNNs}:
  Convolutional layers can exploit absolute spatial location,'' in \emph{Proc.
  IEEE/CVF Conf. Computer Vision and Pattern Recognition (CVPR)}, 2020, pp.
  14\,274--14\,285.

\bibitem{geirhos2020shortcut}
R.~Geirhos, J.-H. Jacobsen, C.~Michaelis, R.~Zemel, W.~Brendel, M.~Bethge, and
  F.~A. Wichmann, ``Shortcut learning in deep neural networks,'' \emph{Nature
  Machine Intelligence}, vol.~2, no.~11, pp. 665--673, 2020.

\end{thebibliography}

\end{document}